\documentclass[11pt,letterpaper]{article}
\usepackage[T1]{fontenc}
\usepackage[utf8]{inputenc}
\usepackage[margin=0.75in,top=0.72in,bottom=0.78in]{geometry}
\usepackage{amsmath,graphicx,booktabs,pifont,xcolor,colortbl,multirow,url,placeins}
\usepackage{newtxtext,newtxmath}
\usepackage{microtype}
\usepackage{caption}
\usepackage{enumitem}
\usepackage[pdfencoding=auto,psdextra,hidelinks]{hyperref}

\hypersetup{
  pdftitle={X2-NativeCursor: Native-Token Text Progress Tracking for Incremental-Text Streaming Codec TTS},
  pdfauthor={Zehan Liu, Carl Chen, Rime Wen, Kaiqi Fu, Altman Lin, Shawn Qin, Lights Shi, Roy Gan, Hao Wang, Qian Wang}
}
\setlist[enumerate]{leftmargin=1.7em,itemsep=3pt,topsep=4pt,parsep=0pt}
\definecolor{NCOursBg}{HTML}{DFF0DC}

\newcommand{\ncsecond}[1]{\underline{#1}}              
\newcommand{\ncna}{--}          
\newcommand{\ncunrep}{$/$}      
\newcommand{\ncref}{\itshape ref.}
\newcommand{\ncyes}{\ding{51}}
\newcommand{\ncno}{\ding{55}}

\newcommand{\pmsd}[2]{#1\,$\pm$\,#2}

\makeatletter
\newcommand{\name}[1]{\gdef\NCAuthorNames{#1}}
\newcommand{\address}[1]{\author{\NCAuthorNames\\[6pt]#1}}
\renewcommand{\maketitle}{%
  \begin{center}
    {\LARGE\bfseries\@title\par}
    \vspace{12pt}
    {\normalsize\begin{tabular}{c}\@author\end{tabular}\par}
  \end{center}
  \vspace{8pt}
}
\renewenvironment{abstract}{%
  \begin{list}{}{\setlength{\leftmargin}{1.5em}\setlength{\rightmargin}{1.5em}}
  \item[]\small\noindent\textbf{\abstractname}\par\vspace{3pt}\noindent
}{\end{list}}
\newenvironment{keywords}{\par\small\noindent\textbf{Index Terms---}\ignorespaces}{\par\vspace{5pt}}
\makeatother
\let\NCoriginalbibliography\thebibliography
\renewcommand{\thebibliography}[1]{%
  \NCoriginalbibliography{#1}%
  \setlength{\itemsep}{1.5pt}%
  \setlength{\parsep}{0pt}%
  \interlinepenalty=10000
}
\date{}
  \title{X2-NativeCursor: Native-Token Text Progress Tracking\\
         for Incremental-Text Streaming Codec TTS}
\name{\shortstack{Zehan Liu, Carl Chen, Rime Wen, Kaiqi Fu, Altman Lin\\
      Shawn Qin, Lights Shi, Roy Gan, Hao Wang, Qian Wang}}
\address{X Square Robot}

\begin{document}
\maketitle

\begin{abstract}
Incremental-text streaming text-to-speech (TTS) needs online text progress tracking for synchronized highlighting, interruption handling, and dialogue-history updates. Input text arrives before it is spoken, so text arrival alone cannot indicate speech progress. Existing waveform-based alignment requires complete audio or adds acoustic processing during streaming. We propose X2-NativeCursor, a lightweight observer that tracks progress from native speech tokens before waveform decoding without changing the TTS generator. Its normalization plan links spoken labels to their original-text spans. Text and native-token encoders feed a local matcher that estimates the current label position. A separate output rule converts revisable position estimates into a cursor that never moves backward. Mean absolute error against an automatic reference is 0.151 Chinese characters with 80-ms lookahead, versus 1.253 characters with 320-ms lookahead for an online waveform baseline. Alignment real-time factor also decreases from 0.3598 to 0.0180 relative to this baseline. Lower tracking error is retained under a second automatic alignment reference. We evaluate X2-NativeCursor on Qwen3-TTS and validate its adaptation to CosyVoice2 by training a separate observer for each backbone. Code is publicly available at \url{https://github.com/X-Square-Robot/X2Streaming-TTS}.
\end{abstract}


\section{Introduction}

Incremental-text streaming text-to-speech (TTS) speaks before a language model finishes generating text \cite{ma2020incremental,stephenson2020lookahead}. Text highlighting, interruption handling, and dialogue-history updates require knowing which words have been spoken \cite{defossez2024moshi}. Text arrival does not specify when individual words are spoken. In codec-based TTS, words span varying numbers of speech-token frames, so frame counts alone cannot locate the current word. Tracking therefore requires an online mapping from speech frames to original-text positions.

Existing methods use waveform-based alignment or alignment built into the TTS generator. Waveform-based aligners estimate timestamps through hidden Markov models (HMMs), connectionist temporal classification (CTC), or timestamp prediction \cite{mcauliffe2026mfa,graves2006ctc,kurzinger2020ctcsegmentation,pratap2024mms,gao2023funasr,shi2023timestamp,rehman2025bfa}. LLM-ForcedAligner predicts timestamps at selected text positions in one non-autoregressive pass \cite{mu2026llmforcedaligner}, while streaming reading trackers estimate positions from incoming speech \cite{sunder2023readingtracker}. Within TTS, Speech-T learns alignment through a transducer \cite{chen2021speecht}. ELLA-V and CTC-TTS arrange text or phonemes with corresponding speech tokens \cite{song2025ellav,liu2026ctctts}; VoXtream generates audio tokens in input-phoneme order \cite{torgashov2025voxtream}.

Complete-audio aligners cannot update progress during generation. Online waveform-based alignment still waits for the TTS waveform decoder to reconstruct audio from speech tokens, then processes it with a separate acoustic encoder. For CTC alignment, the aligner predicts frame-level label probabilities to align audio with known text. This path adds waiting and acoustic computation. Alignment built into a generator may require changing its architecture and retraining it. Native speech tokens already carry speech content and timing \cite{hu2026qwen3tts,du2024cosyvoice2}, allowing tracking before waveform decoding without re-encoding decoded audio. We seek to track growing text and token streams with little computation and bounded lookahead.

We propose X2-NativeCursor, a lightweight online observer operating without waveform input. It reads native speech-token frames, the speech-side representations emitted at successive TTS time steps. A normalization plan converts text into spoken labels describing its normalized reading. Labels are released only when their reading is fixed, so later input cannot change them. They retain their original-text spans, including when numbers or symbols expand into multiple labels. A text encoder represents these labels, while a native-token encoder extracts speech features with bounded lookahead.

For each frame, a local matcher scores labels near the previous estimate using text and speech features. This local search keeps the number of comparisons small as text grows. Internal estimates can move backward or skip labels; an output rule maps the furthest label reached so far back to the original text. The published cursor therefore never retreats. We train a separate observer for each backbone, keeping the TTS generator, speech tokenizer, and waveform decoder unchanged.

Our contributions are as follows:
\begin{enumerate}
\item We introduce an independently trained observer for online text tracking directly from native speech tokens, without waveform input or generator changes.
\item We separate revisable alignment estimates from monotonic cursor output in the original text, with bounded lookahead.
\item Against an automatic reference, Chinese-character mean absolute error (MAE) is 0.151 at 80-ms lookahead, versus the online waveform baseline's 1.253 at 320 ms. Our observer has 2.166M parameters and an alignment real-time factor of 0.0180, versus 0.3598 for this baseline. The tracking gain holds under a second automatic reference. The design supports codec-based TTS models through observer retraining, as demonstrated on Qwen3-TTS and CosyVoice2.
\end{enumerate}

\section{Method}
\label{sec:method}
\begin{figure}[!htbp]
  \centering
  \includegraphics[width=\textwidth]{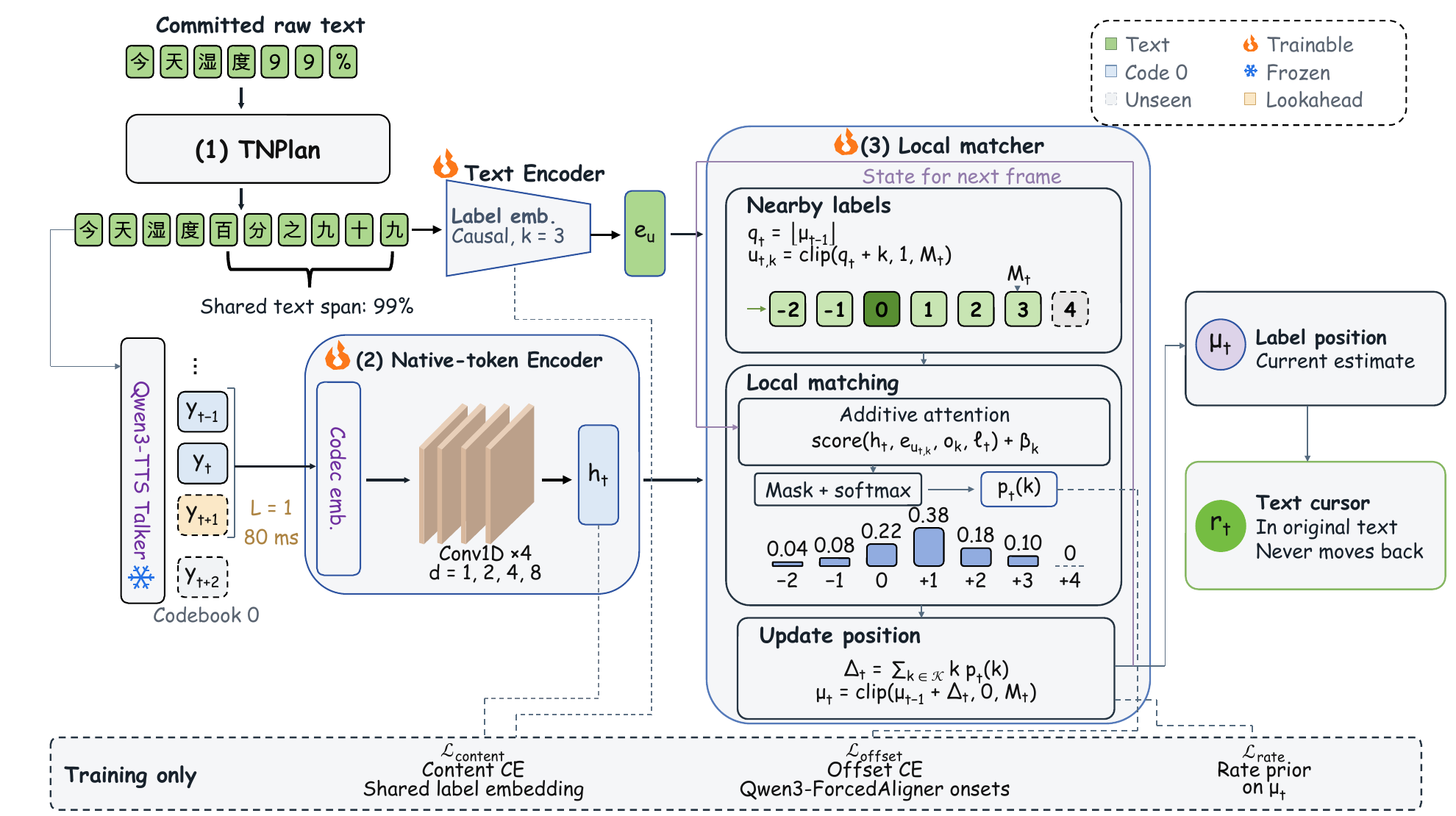}
  \caption{Overview of X2-NativeCursor. TNPlan (1) maps spoken labels to original-text spans. The native-token encoder (2) and local matcher (3) track the current label position before waveform decoding. The output is a cursor in the original text that never moves backward. Dashed paths are used only during training.}
  \label{fig:method}
\end{figure}

\subsection{Overall architecture}
X2-NativeCursor combines text preparation, a trainable observer, and cursor output (Figure~\ref{fig:method}). TNPlan (1) converts incoming text into stable spoken labels and records their original-text spans. The observer comprises the text encoder, native-token encoder (2), and local matcher (3). The text encoder represents labels, while the native-token encoder processes speech-token frames with bounded lookahead. The matcher uses these features to update an internal label position $\mu_t$, which may move backward or skip labels. A separate output rule maps the furthest label position reached to an original-text cursor $r_t$ that never retreats. Only the observer is trained; the TTS generator, speech tokenizer, and waveform decoder remain unchanged.

\subsection{Spoken labels and text mapping}

TNPlan is the normalization plan that converts incoming text into spoken labels and maps them to the original text. It releases labels to the TTS generator and observer once the spoken form is fixed. If later input may change the reading of a number, unit, or symbol, TNPlan waits before releasing its labels. In Chinese, the spoken form of ``99\%'' places the percent expression before the number. All labels for this expression share the original-text span of ``99\%''.

At native frame $t$, the committed text prefix is $x_{1:n(t)}$, containing $n(t)$ original-text characters. The available spoken labels are $z_1,\ldots,z_{M_t}$, where each label $z_u$ is linked to an original-text span $[a_u,b_u)$. A label embedding and a causal convolution with kernel size 3 produce the text features $e_u=\operatorname{TextEnc}(z_{\leq u})$.

\subsection{Native-token encoding}

The native-token encoder maps each speech token $y_t$ to an embedding. Four convolution blocks use dilation factors of 1, 2, 4, and 8. They produce the feature for frame $t$:
\begin{equation}
h_t=\operatorname{AudioEnc}(y_{t-P:t+L}),
\end{equation}
where $P$ is the number of past frames and $L\in\{0,\ldots,4\}$ is the number of future frames. We fix $L$ during training and inference. The main Qwen3-TTS setting uses $L=1$, which corresponds to 80~ms of lookahead. The feature $h_t$ is computed when token $y_{t+L}$ arrives, and the resulting position estimate refers to frame $t$.

\subsection{Local matching and cursor updates}

For each native frame, the matcher scores labels near the previous internal position $\mu_{t-1}$. We use seven offsets $\mathcal K=\{-2,\ldots,4\}$ around $q_t=\lfloor\mu_{t-1}\rfloor$. The label lookup index is $u_{t,k}=\operatorname{clip}(q_t+k,1,M_t)$. Each score combines native-token features, text features, and a location state:
\begin{equation}
s_{t,k}=v^{\top}\tanh\bigl(W_hh_t+W_ee_{u_{t,k}}+W_\ell\ell_t+o_k\bigr)+\beta_k.
\end{equation}
The weights $W_h$, $W_e$, $W_\ell$, and $v$ are learned, with separate parameters $o_k$ and $\beta_k$ for each offset. The state $\ell_t$ contains the fractional part of $\mu_{t-1}$, scaled dwell time, and mean recent advance, all computed before frame $t$. Dwell time counts frames since the label cursor last advanced, scaled by the prior label rate.

We mask candidates with $q_t+k$ outside $[1,M_t]$ and apply a softmax to obtain the offset distribution $p_t(k)$. Its mean gives the position update:
\begin{equation}
\Delta_t=\sum_{k\in\mathcal K}k\,p_t(k),\qquad
\mu_t=\operatorname{clip}(\mu_{t-1}+\Delta_t,0,M_t).
\end{equation}

The published label cursor $c_t$ keeps the furthest position reached. The original-text cursor $r_t$ is the largest span end among labels up to $c_t$:
\begin{equation}
c_t=\max_{s\leq t}\lfloor\mu_s\rfloor,\qquad
r_t=\max_{u\leq c_t}b_u.
\end{equation}
The initial states are $\mu_0=c_0=r_0=0$ and $\ell_1=0$. Before the first label is reached, $r_t$ remains zero.

\subsection{Training and inference}

Qwen3-ForcedAligner provides onset times for the spoken labels \cite{shi2026qwen3asr,mu2026llmforcedaligner}. These times define the reference label position at each native frame. We train the offset distribution $p_t(k)$ with cross entropy against the reference offsets. A content loss predicts the spoken label from each native-token feature $h_t$, using a classifier that shares the label embedding. A rate loss keeps the overall cursor advance close to a prior rate for each utterance. The total loss is
\begin{equation}
\mathcal L=\mathcal L_{\mathrm{offset}}+0.3\mathcal L_{\mathrm{content}}+0.01\mathcal L_{\mathrm{rate}}.
\end{equation}

The encoders use a hidden dimension of 256. We train for 10 epochs with AdamW, a learning rate of $10^{-3}$, cosine decay, and a batch size of 24. During training, we perturb the reference cursor position with probability 0.3. Only the text encoder, native-token encoder, and local matcher are trained. For each TTS backbone, we initialize and train a separate observer to match its token vocabulary and frame period.

During inference, each new token $y_{t+L}$ triggers the update for frame $t$. The observer outputs $c_t$ and $r_t$ and prepares the location state for the next frame.

\section{Experimental Evaluation}
\begin{table}[!htbp]
\centering
\caption{Cursor tracking on Qwen3-TTS speech, with Qwen3-ForcedAligner as the automatic reference. MAE$_{\mathrm{zh}}$ covers the three Chinese groups; MAE$_{\mathrm{en}}$ covers English. $L$: lookahead. $^{\ast}$MMS-FA reference on 600 Chinese and four mixed-language samples, with timestamps shifted 40~ms earlier. $^{\dagger}$Shared MMS-FA emissions. $/$: not evaluated. Bold/underline: best/second best.}
\label{tab:main}
\small
\setlength{\tabcolsep}{2.8pt}
\renewcommand{\arraystretch}{1.12}
\begin{tabular}{c@{\hspace{3pt}}lccrcccrrr}
\toprule
& & \multicolumn{3}{c}{Setting} & \multicolumn{3}{c}{MAE (characters)$\downarrow$} & & \multicolumn{2}{c}{Onset (ms)} \\
\cmidrule(lr){3-5}\cmidrule(lr){6-8}\cmidrule(lr){10-11}
& Method & Online & $L$ (ms) & Params & zh & zh$^{\ast}$ & en & F1@80$\uparrow$ & AAS$\downarrow$ & Lag$_{50}$ \\
\midrule
\multirow{5}{*}{\rotatebox[origin=c]{90}{\itshape Waveform}}
& Qwen3-ForcedAligner-0.6B & \ncno & $\infty$ & 912.6\,M & \ncref & 0.158 & \ncna & \ncna & \ncna & \ncna \\
& MMS-FA & \ncno & $\infty$ & 315.5\,M & 0.539 & \ncref & 1.610 & 0.430 & 109.3 & 80 \\
& CTC-segmentation & \ncno & $\infty$ & 315.5\,M\rlap{$^{\dagger}$} & \ncsecond{0.371} & \ncsecond{0.250} & \ncunrep & 0.418 & 261.7 & 80 \\
& FunASR timestamp predictor & \ncno & $\infty$ & 39.6\,M & 0.744 & 0.669 & \ncunrep & 0.322 & 216.1 & 80 \\
& WindowMMS+PersistentCTC & \ncyes & 320 & 315.5\,M & 1.253 & 1.134 & 2.226 & 0.341 & 172.2 & 80 \\
\addlinespace[2pt]
\multirow{3}{*}{\rotatebox[origin=c]{90}{\itshape Codec}}
& Rate prior & \ncyes & 0 & 0 & 1.706 & 1.750 & \ncunrep & 0.261 & 532.5 & $-240$ \\
& Cross-attention readout & \ncyes & 320 & 2.490\,M & 0.414 & 0.311 & 1.629 & \ncsecond{0.828} & 96.1 & 80 \\
& CodecCTC+skip-DP & \ncyes & 320 & 1.705\,M & 0.416 & 0.311 & \ncsecond{1.568} & 0.823 & \ncsecond{95.6} & 80 \\
\midrule
\rowcolor{NCOursBg}
& \textbf{X2-NativeCursor (ours)} & \ncyes & 80 & 2.166\,M & \textbf{0.151 $\pm$ 0.005} & \textbf{0.206} & \textbf{1.247} & \textbf{0.924} & \textbf{51.1} & \textbf{0} \\
\bottomrule
\end{tabular}
\end{table}

\subsection{Experimental setup}
\label{sec:protocol}

\noindent\textbf{Data.} We train the observer on 20,235 speech samples generated by Qwen3-TTS and evaluate it on 800 fixed test texts. The test set has four groups of 200 texts: plain Chinese, Chinese with numbers, Chinese with symbols, and English.

\noindent\textbf{Streaming setup.} We provide text in chunks of 2--8 characters to simulate incremental input from an LLM. All methods are evaluated on the same test samples. Online methods follow the same text-arrival schedule. We keep the TTS generator unchanged and use a fixed voice. X2-NativeCursor uses one future native frame, corresponding to 80~ms of lookahead. We train the observer with seeds 0, 1, and 2.

\noindent\textbf{References and baselines.} Qwen3-ForcedAligner provides automatic timestamps for observer training and the main evaluation \cite{shi2026qwen3asr,mu2026llmforcedaligner}. The scores measure agreement with this automatic reference. Complete-audio baselines include MMS-FA, CTC-segmentation, and the FunASR/Paraformer timestamp predictor \cite{pratap2024mms,graves2006ctc,kurzinger2020ctcsegmentation,gao2023funasr,shi2023timestamp}. The online waveform baseline, WindowMMS+\allowbreak PersistentCTC, re-encodes a one-second left-context window with at most 320~ms of lookahead. It keeps the CTC alignment state across updates and advances it only with complete encoder outputs. Native-token baselines include a rate prior, a cross-attention readout, and CodecCTC+skip-DP. The latter two use the same alignment supervision as our observer. Table~\ref{tab:main} lists each method's online status and lookahead.

\noindent\textbf{Metrics.} Our main metric is mean absolute error (MAE), which measures the difference between the reported and reference cursor positions in original-text characters. We first average errors within each speech sample, then across samples. We score only frames whose reference positions fall within the committed spoken labels. These frames account for over 99\% of Mandarin frames. English labels map to whole words averaging 4.58 characters, so position updates are coarser in English than in Chinese. A label's onset is the first frame when the published cursor reaches it. We report onset F1 with an 80-ms tolerance and median signed onset lag. We also report onset-only accumulated averaging shift (AAS), the mean absolute difference between predicted and reference label start times \cite{mu2026llmforcedaligner}. We report lookahead separately from real-time factor (RTF), the ratio of computation time to speech duration.

Results from three seeds are reported as the mean and standard deviation.

\subsection{Main results}

Table~\ref{tab:main} compares X2-NativeCursor with waveform-based and native-token baselines on speech generated by Qwen3-TTS. With Qwen3-ForcedAligner as the reference, our method achieves a Chinese-character MAE of 0.151 $\pm$ 0.005. Compared with the online waveform baseline, Chinese and English MAE are lower by about 88\% and 44\%, respectively. This gain is achieved with 80-ms lookahead, compared with 320~ms for the baseline. Chinese MAE is also lower than that of all complete-audio baselines in the table.

The gain also holds among methods that use native tokens. Cross-attention readout and CodecCTC+skip-DP use the same alignment supervision as our observer. Our method reduces Chinese MAE by about 64\% compared with both methods, while using less lookahead.

To check whether the gain depends on the training reference, we re-score 604 aligned samples using MMS-FA, which is not used in observer training. We shift its timestamps 40~ms earlier to account for the median difference between the two references. X2-NativeCursor remains ahead of the online waveform baseline and the two learned native-token baselines, with a Chinese-character MAE of 0.206. The improvement therefore holds under both automatic references.

\subsection{Ablation study}
\label{sec:ablation}
\begin{table}[!htbp]
\centering
\caption{Ablations on Chinese and English samples using offline replay. Lookahead is 80~ms unless removed. MAE averages frame errors within each text group, then across the four groups. Ablations report three-seed means ($\pm$ std); the final row tests new voices.}
\label{tab:ablation}
\small
\setlength{\tabcolsep}{6pt}
\renewcommand{\arraystretch}{1.12}
\begin{tabular}{@{}llcccc@{}}
\toprule
& & \multicolumn{4}{c}{MAE by text subset$\downarrow$} \\
\cmidrule(lr){3-6}
Configuration & MAE$\downarrow$ & zh-plain & zh-num & zh-sym & en \\
\midrule
\rowcolor{NCOursBg}
\textbf{X2-NativeCursor} & \textbf{\pmsd{0.343}{0.009}} & \ncsecond{0.131} & \ncsecond{0.146} & \ncsecond{0.170} & \textbf{0.927} \\
w/o text encoder & \pmsd{1.465}{0.032} & 1.695 & 0.989 & 0.977 & 2.197 \\
w/o backward \& skip & \pmsd{2.681}{0.047} & \textbf{0.124} & 0.155 & \textbf{0.168} & 10.278 \\
w/o lookahead & \pmsd{0.471}{0.026} & 0.188 & 0.197 & 0.250 & 1.251 \\
w/o position/rate & \ncsecond{\pmsd{0.355}{0.023}} & 0.136 & \textbf{0.145} & 0.180 & \ncsecond{0.960} \\
\addlinespace[2pt]
Voice replacement & 0.788 & 0.768 & 1.038 & 0.561 & 0.786 \\
\bottomrule
\end{tabular}
\end{table}

Table~\ref{tab:ablation} reports ablation results on Chinese and English test samples. Removing the text encoder raises MAE from 0.343 to 1.465 and worsens all four groups. Removing the position and rate features has a much smaller effect, with MAE rising to 0.355. Removing backward and skip moves together mainly affects English, where MAE rises from 0.927 to 10.278, while the Chinese results change little. Removing lookahead raises MAE to 0.471. In a separate comparison using seed 0, increasing lookahead from 80 to 160~ms gives no further improvement. Increasing it to 320~ms lowers MAE from 0.333 to 0.310. We therefore use 80~ms to keep lookahead short while retaining most of the accuracy gain. With three voices not used in training, average MAE rises to 0.788, with large differences between voices (0.320--1.213). The observer is therefore sensitive to changes in the synthesis voice.

\subsection{Runtime cost and concurrency}
\label{sec:runtime}

X2-NativeCursor has an RTF of 0.0180, compared with 0.3598 for WindowMMS+\allowbreak PersistentCTC, reducing alignment computation time by about 95\%. In a standalone timing test, the median observer cost is 1.33~ms per native frame, measured after warm-up and excluding waveform decoding. Together with the main results, this shows that the observer lowers both tracking error and computation cost.

\begin{figure}[!htbp]
\centering
\includegraphics[width=0.70\linewidth]{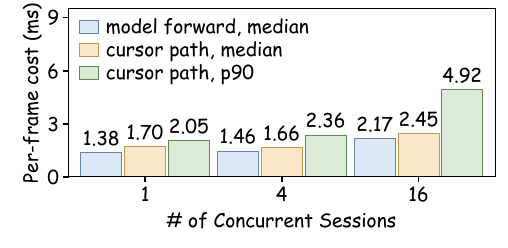}
\caption{Per-frame computation time in the streaming engine at different concurrency levels. Bars show median model-forward time and the median and 90th percentile for a complete cursor update.}
\label{fig:concurrency}
\end{figure}

Figure~\ref{fig:concurrency} reports concurrency results on a single NVIDIA A800-SXM4-80GB GPU. As the number of concurrent sessions increases from 1 to 16, the median time per cursor update rises from 1.70 to 2.45~ms. At 16 sessions, the 90th percentile is 4.92~ms, well below the 80~ms of speech represented by one native frame.

\subsection{Adaptation to other codec-based TTS models}

X2-NativeCursor can be adapted to other codec-based TTS models by retraining the observer on their native speech tokens. We use CosyVoice2 \cite{du2024cosyvoice2} as a test case, reusing the observer architecture and keeping the TTS generator unchanged. The run selected by development MAE achieves a Chinese-character MAE of 0.284 (95\% CI: 0.237--0.343). The other two training seeds give MAEs of 0.254 and 0.241. These results support using the same observer design across codec-based TTS backbones, with retraining for each model.

\begin{figure}[!htbp]
\centering
\includegraphics[width=0.72\linewidth]{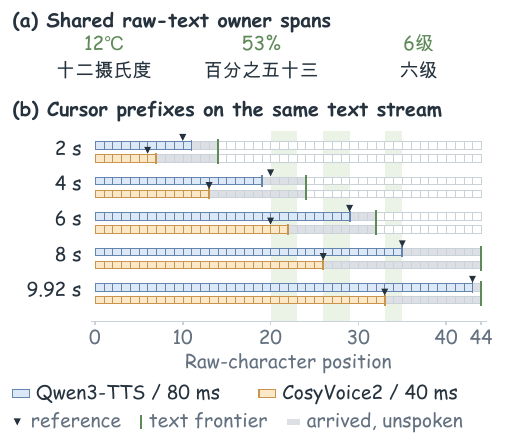}
\caption{Cursor tracking on Qwen3-TTS and CosyVoice2 against their automatic references. (a) Text fields and spoken forms. (b) Cursor positions under the same text input. Each cell represents one original-text character. Native frame periods are 80 and 40~ms, respectively.}
\label{fig:backbone}
\end{figure}

Figure~\ref{fig:backbone} shows tracking for the same sentence under a shared text-input schedule. The two TTS systems speak at different rates in this example. Across the ten plotted estimates, each cursor stays within two characters of its own automatic reference and within the text received so far. This example shows how the observers follow the speech progress of their respective backbones.

\FloatBarrier

\section{Conclusion}
X2-NativeCursor tracks raw-text progress through local native-token state updates and owner-span projection, while keeping the TTS frozen. On Qwen3-TTS, it reduces teacher-relative Chinese-character MAE from the online waveform baseline's 1.253 to 0.151 with one-quarter of the lookahead. The improvement also holds with a second automatic reference, and retraining the observer adapts the method to CosyVoice2. These results support online cursor estimation before waveform decoding within the evaluated conditions.

\section{Compliance with Ethical Standards}
This study uses synthetic speech produced by the systems under test and
includes no human participants.

\section{Acknowledgments}
This work was supported by X Square Robot, which provided funding and
computational resources. All authors are employees of X Square Robot.

\clearpage
\begingroup
\fontsize{9.5}{11.3}\selectfont
\setlength{\parskip}{0pt}
\bibliographystyle{IEEEbib}
\bibliography{refs}
\endgroup

\end{document}